\documentclass[letterpaper, 10 pt, conference]{ieeeconf}  % Comment this line out if you need a4paper

\IEEEoverridecommandlockouts                              % This command is only needed if 
\usepackage{cite}
\usepackage{amsmath,amssymb,amsfonts}
\usepackage{algorithmic}
\usepackage{graphicx}
\usepackage{textcomp}
\usepackage{xcolor}
\usepackage{epsfig}
\usepackage{hyperref}

\title{\LARGE \bf
Fast Hierarchical Neural Network for Feature Learning on Point Cloud
}

\author{Can Chen$^{1}$,  Luca Zanotti Fragonara$^{2}$ and Antonios Tsourdos$^{3}$% <-this % stops a space
}

\begin{document}

\maketitle
\thispagestyle{empty}
\pagestyle{empty}

%%%%%%%%%%%%%%%%%%%%%%%%%%%%%%%%%%%%%%%%%%%%%%%%%%%%%%%%%%%%%%%%%%%%%%%%%%%%%%%%
\begin{abstract}
The analyses relying on 3D point clouds are an utterly complex task, often involving million of points, but also requiring computationally efficient algorithms because of many real-time applications; e.g. autonomous vehicle. However, point clouds are intrinsically irregular and the points are sparsely distributed in a non-Euclidean space, which normally requires point-wise processing to achieve high performances. Although shared filter matrices and pooling layers in convolutional neural networks (CNNs) are capable of reducing the dimensionality of the problem and extracting high-level information simultaneously, grids and highly regular data format are required as input. In order to balance model performance and complexity, we introduce a novel neural network architecture exploiting local features from a manually subsampled point set. In our network, a recursive farthest point sampling method is firstly applied to efficiently cover the entire point set. Successively, we employ the k-nearest neighbours (knn) algorithm to gather local neighbourhood for each group of the subsampled points. Finally, a multiple layer perceptron (MLP) is applied on the subsampled points and edges that connect corresponding point and neighbours to extract local features. The architecture has been tested for both shape classification and segmentation using the ModelNet40 and ShapeNet part datasets, in order to show that the network achieves the best trade-off in terms of competitive performance when compared to other state-of-the-art algorithms.
\end{abstract}

%%%%%%%%%%%%%%%%%%%%%%%%%%%%%%%%%%%%%%%%%%%%%%%%%%%%%%%%%%%%%%%%%%%%%%%%%%%%%%%%
\section{INTRODUCTION}

The widespread use of laser scanning and similar 3D sensing technologies rendered 3D point clouds a fundamental format for 3D geometric data. This format is increasingly popular and widely applied to many applications, such as autonomous vehicle \cite{zhou2018voxelnet, qi2018frustum, ku2018joint, liu2018real}, robotic mapping and navigation \cite{biswas2012depth, zhu2017target}, 3D shape representation and modelling \cite{golovinskiy2009shape}. Among the feature-learning techniques, convolutional neural networks (CNNs) show efficiency and significant success in machine vision tasks such as object classification, detection and semantic segmentation. CNNs are easily capable to learn local features from images or videos, as they all have a fixed-sized grid structure. Deep learning operations, such as convolution, pooling, can be readily applied on these sort of structured data. However, CNNs cannot be applied directly to 3D point set due to their irregular, unordered structure. As a result, in order to fully exploit geometric shapes from 3D point sets in an efficient way remains an open challenge, especially when large amount of points are to be processed such as in Lidar sensing for autonomous vehicle.

As a pioneering approach, PointNet \cite{qi2017pointnet} resolved the challenging problem of applying deep learning on unordered 3D point sets. In fact, PointNet exploits point-wise features by multiple layer perceptron (MLP) and gathers all the individual features together to generate a global feature by a pooling layer. Besides, PointNet also provide a theoretical analysis that a symmetric function can be used to aggregate feature from unordered points. However, PointNet only learned global features and limited in capturing local contextual information. In order to address this problem, many researchers investigate effective ways to aggregate local features. PointNet++ \cite{qi2017pointnet++} introduced a hierarchical neural network in order to apply PointNet with designed sampling and grouping layers applied recursively to capture local features from multi-scale local neighbourhood. DGCNN \cite{wang2018dynamic} exploits local contextual information by applying edge convolution on points and corresponding edges that connects to neighbours.

Our proposed Fast Hierarchical neural network for feature learning on point clouds (FastPointNN) has to address two problems: how to subsample in an efficient way from the whole 3D point set and how to extract local features with minimum information loss. Inspired from the solutions of PointNet++ and DGCNN, we designed an effective neural network achieving high performance and low complexity for object classification and semantic part segmentation on 3D point clouds. The key contributions of our work are summarized as follows:
\begin{itemize}
\item We utilize the farthest point sampling method iteratively in order to subsample 3D point set and use $k$-nearest neighbours ($k$-nn) to gather $k$ neighbours. The number of $k$ varies depending on the number of points in the subset. Besides, in order to reduce the sampling loss, the neighbours of the subset are selected from the points before sampling, which means that the information of the abandoned points is kept in the neighbourhood.
\item We employ edge convolution to extract local features for 3D point subsets. The feature of each point consist of self-feature and edges between point and corresponding neighbours.
\end{itemize}

\section{Related Work}

\subsection{Learning features from volumetric Grid}

In order to take full advantage of standard CNNs operations on irregular and sparse 3D point set, it is intuitive to voxelize the point cloud to sparse and uniform 3D grid. \cite{maturana2015voxnet} rasterizes the point cloud into $32^{3}$ voxels with binary state indicating whether the voxel is occupied or not. 3D Convolutional Network Layers are then applied for object classification starting from RGB-D data. However, these 3D dense and gridded voxel data require large memory and computational effort due to the sparse structure of the 3D data. Some improvements with respect to the sparsity problem of the volumetric representation have been proposed by Wang et al \cite{wang2015voting}, who achieved a higher efficiency by using a voting scheme from only the occupied voxels for object detection from a rasterized point cloud. Another concept different from a uniform 3D grid is proposed by Kd-Net \cite{klokov2017escape}, which uses a kd-tree structure  \cite{bentley1975multidimensional} to form the computational graph, based on which, multiplicative transformations are performed according to the subdivisions of the point clouds. OctNet \cite{riegler2017octnet} exploits the sparsity of the regular grid by hierarchically building a partition of the space that generated from a set of unbalanced \textit{octrees} where each leaf node stores a pooled feature representation.

\subsection{Learning features from unstructured point cloud directly}

PointNet \cite{qi2017pointnet} proved possible that the features of an unordered point cloud can be captured using a multiple layer perceptron (MLP) network and a symmetric function like max pooling. Experimentally, PointNet applies the MLP operation on each point individually to extract the corresponding features, followed by a max pooling layer used to summarize the global features for the whole point cloud. This approach achieved state-of-the-art performance on point cloud understanding tasks. However, the architecture has limited capabilities of capturing local features from local regions. Some attempts were made to address this problem by involving local features in hand-crafted or explicit manners. For instance, PointNet++ \cite{qi2017pointnet++} introduces a hierarchical neural network combining the PointNet layer, a sampling layer and a grouping layer recursively. In details, the sampling layer utilizes the farthest point sampling (FPS) method to obtain subsets that can cover the whole point cloud, the grouping layer clusters multi-scale neighbours in a pyramid-like way to construct local regions. The PointNet layer encodes local regions to feature vectors. The enhanced architecture boosted performance significantly when compared with the previous PointNet architecture. DGCNN \cite{wang2018dynamic} constructs local regions by building k-nearest neighbours (kNN) graphs for each point, and encodes each point features by aggregating each point and corresponding edge connecting to the neighboring pairs. Convolution-like operation with a max pooling layer is then applied to extract local features. In order to apply standard CNN operation to an irregular 3D point set, PointCNN \cite{li2018pointcnn} try to learn a $\chi$-convolutional operator to transform a given unordered point set to a latent canonical order. Similar with typical CNN, a hierarchical neural network based on a $\chi$-convolutional operator is constructed for local features extraction.

\subsection{Learning features from RNN-based models}
Inspired by the attention mechanism and sequence to sequence model \cite{bahdanau2014neural}, Point2Sequence \cite{liu2018point2sequence} proposes a RNN-based architecture to encode local features by capturing existing correlation among multi-scale areas with attention. Specifically, farthest point sampling (FPS) is firstly adopted to select points and establish multi-scale areas as local regions. Then, the features of local regions are captured by a shared MLP layer. At last, an RNN-based sequential encoder-decoder model is applied to gather features of all local regions. Meanwhile, an attention mechanism is employed to highlight the importance of each local region.

\subsection{Learning features from multi-view models}
Encouraged by the great success of CNN on vision-based tasks, multi-view based approaches \cite{qi2016volumetric, wang2017dominant} project 3D point cloud into multiple 2D views and applies typical 2D CNNs to realize the 3D point cloud learning task in an indirect way. This is possible using multi-views, chosen from various perspectives of the object, which are capable to represent different geometric properties. Aggregating features of multi-views that are generated from corresponding CNNs operation also leads to impressive performance for the point cloud processing tasks. However, it is non-trivial to extend these multi-view approaches to the semantic segmentation task for 3D point clouds, which requires classification for each point, as multi-views are only 2D images without any depth information.

\subsection{Learning features from geometric deep learning}
Geometric deep learning \cite{bronstein2017geometric} is a term used to define a set of emerging techniques that attempts to leverage deep neural networks to deal with non-Euclidean structured data, such as 3D point cloud, social networks or genetic networks. More recently, \cite{bruna2013spectral} introduced a spectral graph CNN based on Laplacian operator \cite{shuman2012emerging}. However, the Laplacian operation is computationally expensive because of the Laplacian eigendecomposition. The following-up work \cite{defferrard2016convolutional} addressed this problem by avoiding the Laplacian eigendecomposition operation. In the 3D point cloud domain, PointGCN \cite{zhang2018graph} encodes the spatial local structure into the constructed graph, and applies graph convolution operations to learn local contextual features for the classification task.

\section{Our Approach}
Considering applications on large and complex geometric 3D point cloud data, we propose a fast hierarchical neural network to achieve high performance with less computational complexity for both the 3D point cloud classification and semantic part segmentation tasks.

\begin{figure*}[t!]
  %\vspace{-0.2cm}
  \centering
   {\epsfig{file = 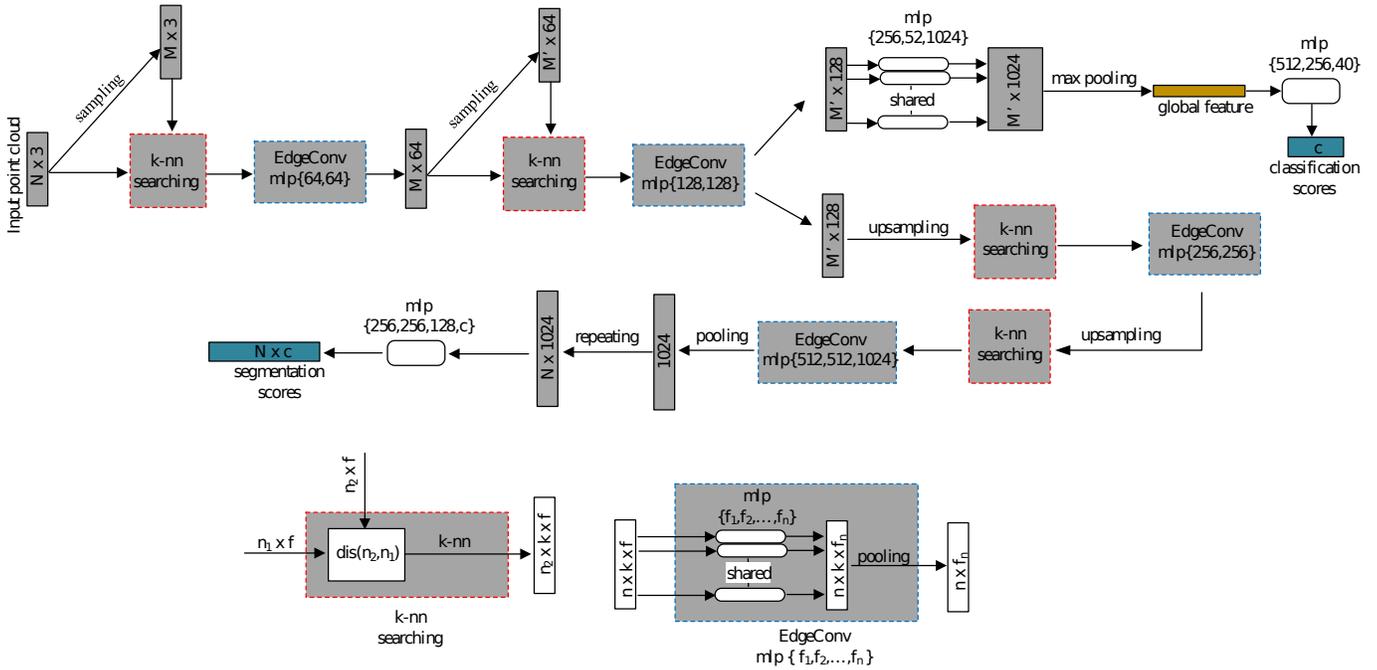, width = 18cm}}
  \caption{\textbf{Model architectures:} The model architecture contains two parts: classification (top branch) and semantic part segmentation (bottom branch). The classification branch takes $n$ unordered points as input, followed by two layers recursively, each of which includes a sampling module, a $k$-nn searching module and an EdgeConv module to aggregate local features from neighbours. Successively, three MLP layers with a max pooling operation are applied to capture the global feature of the whole point cloud. The global feature is finally transformed to 40 categories via shared fully-connected layers for a classification task. The semantic part segmentation model (bottom branch) extends the classification model by combining upsampling model  as shown in Figure \ref{fig:upsampling} for details. \textbf{$k$-nn searching module:} the input tensor with shape $n_1 \times f$ represents $n_1$ points with $f$-dimension before sampling, meanwhile the input tensor with shape $n_2 \times f$ represents $n_2$ points with $f$-dimension after sampling. We calculate the Euclidean distance for each pair and then search $k$ nearest points from $n_1$ for each point in $n_2$. The dimension of the output is $n_2 \times k \times f$. \textbf{EdgeConv module:} the EdgeConv module takes the output tensor of $k$-nn searching module as input, and then applies MLP operations sequentially with the number of neurons defined as $\left\{f_1,f_2,\ldots,f_n\right\}$ At last, a max pooling operation is applied to generate edge features with the shape of $n \times f_n$.}
  \label{fig:architecture}
 \end{figure*}

Figure \ref{fig:architecture} indicates our architecture for details. Suppose that the input to our layers is a given raw 3D point cloud, denoted by $P=\left\{ p_i \in \mathbb{R}^3, i=1,2,\ldots,N\right\}$, where $N$ is the number of the points. We first select $M$ points by iterative farthest point sampling (FPS) as a subset $P1$ from $P$, followed by a $k$-NN processing to find $k$ nearest points in $P$ for each point in $P1$ to aggregate sufficient information for subsampled points.  We then extract a $F$-dimensional feature by multiple edge convolutional operations \cite{wang2018dynamic}. We apply these modules recursively to assemble a hierarchical structure.

\subsection{Sampling Strategy}
We use an iterative farthest point sampling (FPS) strategy to select points for subset. Suppose a empty subset $P1$, a random point is firstly picked and added to $P1$, then it iteratively adds to $P1$ a point that has the farthest distance to the last picked point until expected $M$ points are picked. The FPS strategy has the better coverage of the whole point set than random sampling.

\subsection{Neighbours Searching Strategy}
Being agnostic of the data distribution, the $k$-nearest neighbours ($k$-NN) searching method is used to obtain $k$-neighbours, and the number of $k$ is dynamic and determined by the number of points in the subset. However, considering reduction of spatial resolution during sampling, we embed information of abandoned points to the neighbours. Suppose that $P$ and $P1$ are the point set before sampling and point subset after sampling respectively. We search $k$ nearest points from $P$ for each point in $P1$.

\subsection{Edge Convolution}
Edge Convolution already showed its benefits in extracting local features for 3D point clouds in DGCNN \cite{wang2018dynamic}. Thus, we have chosen to adopt the edge convolution to capture local features of each point associated with edges that connect to neighbours. The operation of edge convolution is defined as Equation \ref{eq:edge_conv}, such that $x'_i$ is update features that associate self feature with corresponding neighbours for each point,  ${\square}$ indicates a generic symmetric function, such as pooling or summation, $x_i$ represents information of each point, which is associated with its neighbours that is regarded as $x_j$. The $h$ is considered as multi-layer perceptron (MLP) to learn features. In summary, the edge convolution operation applies a MLP to each point and corresponding neighbours to capture receptive fields, followed by a max pooling operation to generate a new point set with new dimensional features associated with local features.

\begin{equation}\label{eq:edge_conv}
    x'_i=\underset{j:(i,j)\in\epsilon }{\square}h_\Theta (x_i,x_j-x_i)
\end{equation}

\section{Experiments}
In this section, we evaluate our models for different tasks on a 3D point cloud: specifically, for classification and semantic part segmentation. We compare our complexity, computation and accuracy performance with several state-of-the-art methods, and also visualize experimental results for semantic part segmentation task.

\subsection{Classification}
\textbf{Dataset} We evaluate our classification model on the ModelNet40 benchmark \cite{wu20153d} for 3D object classification. The dataset consists of 12,311 meshed CAD models from 40 man-made object categories. Among the models, 9,843 models are separated for training, and 2,468 models are for testing. Following the data processing of \cite{qi2017pointnet}, we uniformly sample each model to 1,024 points from mesh faces and then normalize them to a unit sphere. The training dataset is also augmented by means of randomly rotating, scaling each object. We further jitter the location of each point for all the objects by a Gaussian noise with zero mean and 0.01 standard deviation.

\textbf{Model Structure} The classification model is presented in Figure \ref{fig:architecture} (top branch). We firstly sample 512 points as a subset by farthest point sampling (FPS) for the local geometric feature extraction in the following layer of edge convolutional operations (64, 64). Similarly, 128 points are then subsampled, followed by two edge convolutional operations (128, 128) as the second layer. The third layer is used to aggregate local features by three MLP layers (256, 512, 1024). For the fourth layer, global feature of point cloud is extracted by a max pooling operation. At last, three fully-connected layers (512, 256, 40) are used to transform the global feature to 40 categories. Dropout operation with a keep probability of 0.5 is also used in the last three fully connected layers. ReLU was used as an activation function and batch normalization is used for all the edge convolution operations and fully-connected layers. Besides, the numbers $k$ of nearest neighbours for the first and second layer are 20 and 15 respectively.

\textbf{Training Details} Same training strategy as DGCNN \cite{wang2018dynamic}, our optimizer is Adam \cite{kingma2014adam} with momentum 0.9, batch size 32 and initial learning rate 0.001 which is divided by 2 every 20 epochs to 0.00001. The decay rate for batch normalization is 0.7 initially and grows to 0.99 gradually. We leverage TensorFlow to train our model on a GTX1080Ti GPU.

\textbf{Results} Table \ref{tab:cls_result} shows our results and the comparison with several state-of-the-art algorithms. Even if we subsampled the point set, our model still achieves competitive and impressive result on the ModelNet40 dataset. The result indicates that, considering large scale of point set in many applications, a well-chosen subset still can achieve acceptable performance for the best trade-off.

\begin{table}[h]
\renewcommand\arraystretch{1.3}
\caption{Classification results on ModelNet40 dataset.}
\label{tab:cls_result} \centering
\begin{tabular}{ccc}

\hline
\multicolumn{1}{}{} & \begin{tabular}[c]{@{}c@{}}MEAN CLASS \\ ACCURACY (\%) \end{tabular} & \begin{tabular}[c]{@{}c@{}}OVERALL \\ ACCURACY  (\%) \end{tabular} \\ \hline
VOXNET                 & 83.0                                                           & 85.9                                                        \\
POINTNET               & 86.0                                                           & 89.2                                                        \\
POINTNET++             & -                                                              & 90.7                                                        \\
KD-NET                 & -                                                              & 91.8                                                        \\
DGCNN                  & 90.2                                                           & 92.2                                                        \\ \hline
OURS                   & 88.1                                                           & 91.1                                                        \\ \hline
\end{tabular}
\end{table}

\textbf{Computational Complexity} We compare the complexity of our classification model to several state-of-the-art by measuring model size, forward time and classification overall accuracy. It is worth noting that, considering different experimental environment, we evaluate again all the models that in the table  \ref{tab:complexity} to align experimental environment. It shows that our model has the least number of parameters and computational complexity, and comparable forward time. Compared with PointNet \cite{qi2017pointnet}, our model outperforms by 1.9\% accuracy, although slower by 1.1ms.

\begin{table}[h]
\renewcommand\arraystretch{1.3}
\caption{Comparation of different models for complexity, forward time and overall accuracy.}
\label{tab:complexity} \centering
\begin{tabular}{cccc}
\hline
\multicolumn{1}{}{} & \begin{tabular}[c]{@{}c@{}}model\\ size (MB)\end{tabular} & \begin{tabular}[c]{@{}c@{}}forward \\ time (MS)\end{tabular} & \begin{tabular}[c]{@{}c@{}}overall \\ accuracy (\%)\end{tabular} \\ \hline
POINTNET               & 41.8                                                      & 14.7                                                        & 89.2                                                            \\
POINTNET++             & 19.9                                                      & 32                                                          & 90.7                                                            \\
DGCNN                  & 22.1                                                      & 52                                                          & 92.2                                                            \\
OURS                   & \textbf{17.1}                                                      & 15.8                                                        & 91.1                                                            \\ \hline
\end{tabular}
\end{table}

\begin{table*}[t!]
\renewcommand\arraystretch{1.3}
\caption{Semantic part segmentation results on ShapeNet part dataset.}
\label{tab:seg_result} \centering
\begin{tabular}{p{1.2cm}|p{0.5cm}|p{0.4cm}p{0.4cm}p{0.4cm}p{0.4cm}p{0.4cm}p{0.4cm}p{0.4cm}p{0.4cm}p{0.4cm}p{0.4cm}p{0.4cm}p{0.4cm}p{0.4cm}p{0.4cm}p{0.4cm}p{0.4cm}}
\hline
                                                        & mean & areo & bag  & cap  & car  & chair & \begin{tabular}[c]{@{}c@{}}ear\\ phone\end{tabular} & guitar & knife & lamp & laptop & motor & mug  & pistol & rocket & \begin{tabular}[c]{@{}c@{}}skate\\ board\end{tabular} & table \\ \hline
\begin{tabular}[c]{@{}c@{}}shapes\\ number\end{tabular} &      & 2690 & 76   & 55   & 898  & 3758  & 69                                                  & 787    & 392   & 1547 & 451    & 202   & 184  & 283    & 66     & 152                                                   & 5271  \\ \hline
pointnet                                                & 83.7 & 83.4 & 78.7 & 82.5 & 74.9 & 89.6  & 73.0                                                & 91.5   & 85.9  & 80.8 & 95.3   & 65.2  & 93.0 & 81.2   & 57.9   & 72.8                                                  & 80.6  \\
pointnet++                                              & 85.1 & 82.4 & 79.0 & 87.7 & 77.3 & 90.8  & 71.8                                                & 91.0   & 85.9  & 83.7 & 95.3   & 71.6  & 94.1 & 81.3   & 58.7   & 76.4                                                  & 82.6  \\
kd-net                                                  & 82.3 & 82.3 & 74.6 & 74.3 & 70.3 & 88.6  & 73.5                                                & 90.2   & 87.2  & 81.0 & 94.9   & 57.4  & 86.7 & 78.1   & 51.8   & 69.9                                                  & 80.3  \\
dgcnn                                                   & 85.1 & 84.2 & 83.7 & 84.4 & 77.1 & 90.9  & 78.5                                                & 91.5   & 87.3  & 82.9 & 96.0   & 67.8  & 93.3 & 82.6   & 59.7   & 75.5                                                  & 82.0  \\ \hline
ours                                                    & 85.0 & 83.7 &\textbf{84.6} & 83.1 & \textbf{78.9} & 90.7  & 72.5                                                & 90.8   & \textbf{87.4}  & \textbf{83.8} & 95.3   & 65.1  & 94.0 & 76.3   & 58.2   & 75.1                                                  & 82.0  \\ \hline
\end{tabular}
\end{table*}

\subsection{Semantic Part Segmentation}
\textbf{Dataset} We further extend our model to adjust for semantic part segmentation task on ShapeNet part dataset \cite{yi2016scalable}. The dataset contains 16,881 3D models from 16 categories, and we sample each 3D model to 2048 points uniformly, each of which is annotated with to a certain one of 50 part classes. Besides, each point set contains several but less than 6 parts. We separate the dataset into training set, validation set, testing set respectively in our experiment.

\textbf{Model Structure} The task of semantic part segmentation is to predict a part category label for each point in the point set, which means that we need to extract features for all the points. Normally, there are two solutions to achieve this task. One is to combine local features and global feature that is duplicated with $N$ times \cite{qi2017pointnet, wang2018dynamic}, which however leads to high amounts of computation. The other solution is to upsample points by interpolation \cite{qi2017pointnet++}. In this paper, considering subsampled point set and required fine-grained information, two methods are used in our structure as shown in Figure \ref{fig:architecture} (bottom branch).

\begin{figure}[h]
  %\vspace{-0.2cm}
  \centering
   {\epsfig{file = 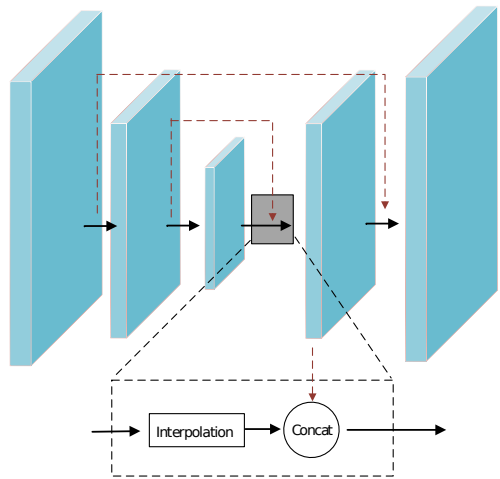, width = 6.5cm}}
  \caption{Upsampling Structure}
  \label{fig:upsampling}
 \end{figure}

In order to generate interpolated points from given subsampling points and corresponding features, we use inverse square Euclidean distance weighted average function based on each point and corresponding $k$ nearest neighbours \cite{qi2017pointnet++} that is shown in Equation \ref{eq:interpolation}, where $\omega_i(x)=\frac{1}{(x-x_i)^2}$ is the inverse square Euclidean distance between $x$ and $x_i$. 

\begin{equation}\label{eq:interpolation}
    f(x)=\frac{\sum_{i=1}^{k}\omega_i(x)f_i }{\sum_{i=1}^{k}\omega_i(x)}
\end{equation}

We then concatenate interpolated points features with corresponding abstraction level point features  as shown in Figure \ref{fig:upsampling}, and apply edge convolution to fuse them together. We borrow first two sampling layers and corresponding edge convolution layers from the classification structure, and obtained the tensor with the shape $128 \times 128$ (points, features). Next, two upsampling layers ($\rightarrow$ 512 $\rightarrow$ 2048) and corresponding edge convolution layers ([256,256], [512,512,1024]) are employed to extract fine-grained features, and the number of neighbours are 15 and 20 respectively. Similar with classification structure, global feature of point cloud is obtained by a max pooling, we then duplicate the global feature with 2048 times and finally apply four multiple layer perceptron (MLP) layers (256,256,128,50) with dropout probability 0.6 to transform the global feature to 50 categories.

\textbf{Training Details} We adopt the same training setting as in classification task, except the batch size of 8, and the training scheme is distributed to two NVIDIA TESLA V100 GPUs, each of which equips 32 GB memory.

\begin{figure}[h]
  %\vspace{-0.2cm}
  \centering
   {\epsfig{file = 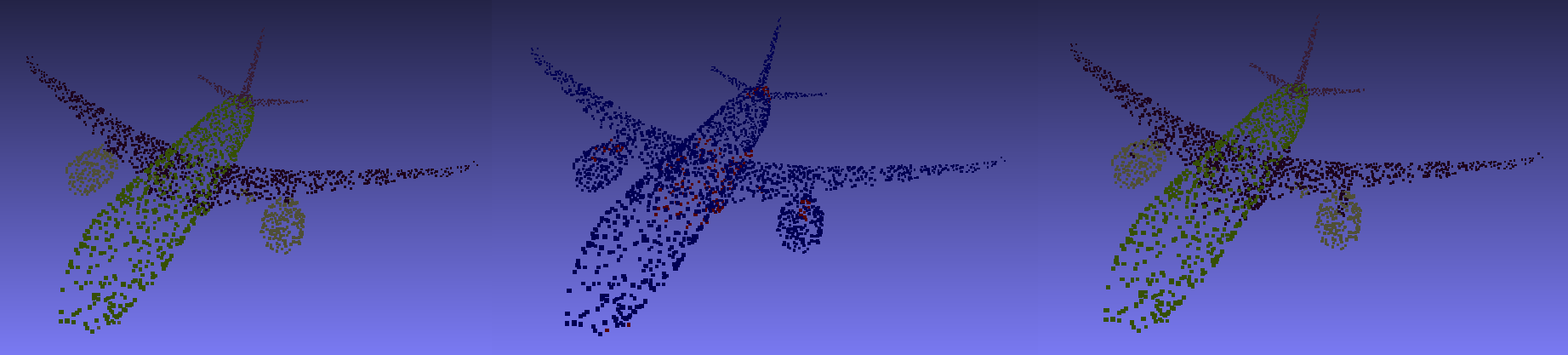, width = 7cm}}
   {\epsfig{file = 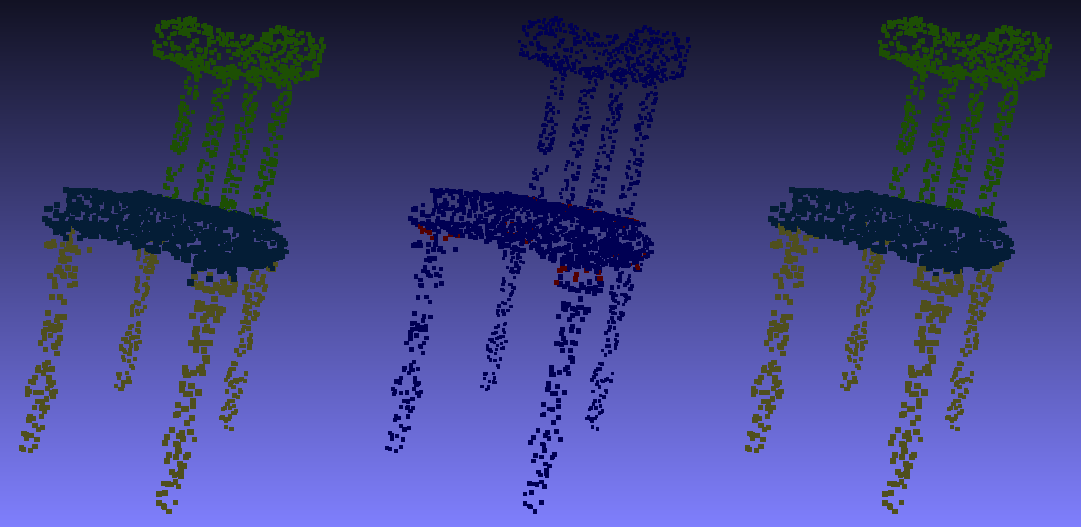, width = 7cm}}
   {\epsfig{file = 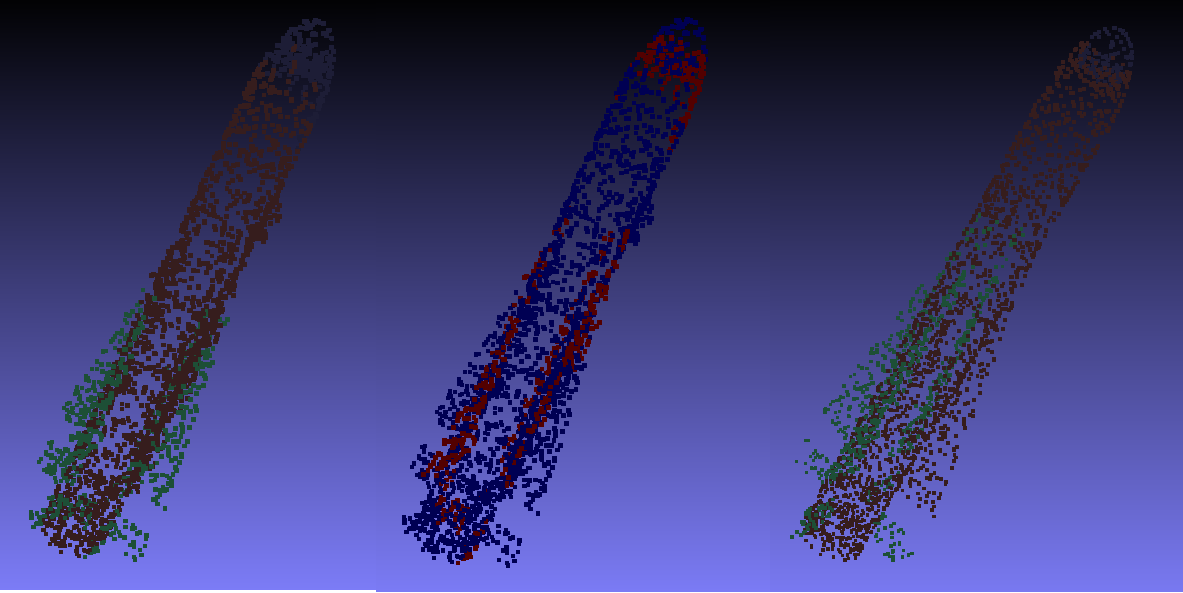, width = 7cm}}
  \caption{Visualization of semantic part segmentation results. For each category, from left to right: prediction result, difference between prediction and ground truth (red color points), ground truth.}
  \label{fig:visual}
 \end{figure}

\textbf{Results} In order to align the evaluation metric, we employ mean Intersection over Union (mIoU) \cite{qi2017pointnet} as our evaluation scheme. The IoU is calculated by difference between ground-truth and prediction for different parts in each shape model, then the mIoU is further obtained by calculating the average of the IoUs for all the shape models int the same category. We compare our model with others as shown in Table \ref{tab:seg_result}, which indicates that our model also achieves competitive results on the ShapeNet part dataset \cite{yi2016scalable}. We also visualize some shapes for our results as shown in Figure \ref{fig:visual}.

\section{CONCLUSIONS}

In this paper, we present an effective and fast hierarchical neural network for feature learning on point cloud and evaluate its complexity and performance on both a classification and semantic part segmentation task using the ModelNet40 and ShapeNet part dataset, respectively. The results shows that our model achieves the best trade-off performance between complexity and accuracy, and obtain similar or even better performance than other state-of-the-art methods. In the future, it seems to be promising to investigate convolution-like operations instead of using multiple layer perceptrons on irregular 3D point clouds.

\section*{ACKNOWLEDGEMENT}

The HumanDrive project is a CCAV $\setminus$ Innovate UK funded R\&D project (Project ref: 103283) led by Nissan and supported by world-class experts from nine other industry and academic organisations (Hitachi, Horiba MIRA, Aimsun, Atkins, University of Leeds, Cranfield University, Highways England, SBD Automotive Ltd. and the Transport Systems Catapult), is developing an advanced vehicle control system, designed to allow the vehicle to emulate a "natural" human driving style using machine learning and developing an Artificial Intelligence to enhance the user comfort, safety and experience. Find out more: \url{www.humandrive.co.uk}

%\bibliographystyle{IEEEtran}
%\bibliography{bibliography}

\end{document}